\documentclass{svproc}
\usepackage{url}
\usepackage{graphicx, hyperref, svg}
\begin{document}
\mainmatter              % start of a contribution
\title{Dataset Generation and Bonobo Classification from Weakly Labelled Videos}
\titlerunning{Dataset Generation and Bonobo Classification}  % abbreviated title (for running head)
\author{Pierre-Etienne Martin}

% \author{Pierre-Etienne Martin\orcidID{0000-0002-9593-4580}}
% \institute{}
%  \authorrunning{P.-E. Martin}
%
\authorrunning{P.-E. Martin} % abbreviated author list (for running head)
% \tocauthor{}
%
\institute{CCP Department, Max Planck Institute for Evolutionary Anthropology, D-04103 Leipzig, Germany\\
\email{pierre\_etienne\_martin@eva.mpg.de}}

\maketitle              % typeset the title of the contribution

\begin{abstract}
This paper presents a bonobo detection and classification pipeline built from the commonly used machine learning methods. Such application is motivated by the need to test bonobos in their enclosure using touch screen devices without human assistance. This work introduces a newly acquired dataset based on bonobo recordings generated semi-automatically. The recordings are weakly labelled and fed to a macaque detector in order to spatially detect the individual present in the video. Handcrafted features coupled with different classification algorithms and deep-learning methods using a ResNet architecture are investigated for bonobo identification. Performance is compared in terms of classification accuracy on the splits of the database using different data separation methods. We demonstrate the importance of data preparation and how a wrong data separation can lead to false good results. Finally, after a meaningful separation of the data, the best classification performance is obtained using a fine-tuned ResNet model and reaches 75\% of accuracy.


\end{abstract}
\section{Introduction}
\label{sec:intro}
Direct application of computer vision tools in a specific domain is rarely possible. Parameters often need to be fine-tuned to adapt to scene changes, model purpose or output size. Most applications often imply real-time processing leading to a trade-off in the conception of the pipeline between model performance, pipeline complexity, hardware capacity and processing time. It is with such a principle that many open source projects, such as Scikit-learn~\cite{ScikitLearn_2011}, DeepLabCut~\cite{Deeplabcut_Mathisetal2018}, Detectron2 from Facebook~\cite{detectron2_facebook_2019} or MMDetection from OpenMMLab~\cite{mmdetection_2019}, came into being. Their goal is to deliver updated computer vision tools in a wide range of applications by providing model architecture and their pre-trained weights on different datasets. Therefore, industry applications and academic researchers are saving time and computation resources by avoiding the training process of such methods. However, for a specific application such as Bonobos classification, fine-tuning seems to remain a necessary step in such a process.
\par
Great-Ape individual classification is not a new topic in the field of behaviour analysis or computer vision. In 2012, three datasets tackling such a problem were presented: Gorillas, C-Zoo and C-Tai~\cite{Primate_id_handcraftedfeatures_2012}. The two latest are then re-used and refined by different teams to improve the state-of-the-art classification methods. In~\cite{ChimpdetectandidentifSURFGABOR_2013}, the author hand-annotated the two mentioned datasets using Image Maker and provided a classification solution based on Gabor features and local Surf descriptor. With the progress of the deep learning methods in the following years, the authors of~\cite{CNNChimpReco_2016} presented a CNN-based model in order to predict different attributes such as identity, age, age group, and gender of chimpanzees. Because of the wide use of face recognition on humans in the same years, transposition to apes and mammals, in general, is investigated~\cite{Animal_biometric_2017}, suggesting face features could be universal. It is with such principle that~\cite{Rhesus_macaques_2018} and~\cite{pandaidentification_2020} address similar classification problems respectively to rhesus macaques and pandas.
\par
With the many introduced dataset for primates, \cite{Animal_Face_Primates_2020}~compiles the Animal Face Dataset gathering 41 primate species from wild and captive animals. By selecting the 17 most populated species in the dataset, they reached 93.6\% of individual classification. Similar work was conducted on Chimps only with CNN~\cite{CNN_Chimp_recognition_2019} on long-term video records in a 14-year dataset (10 million face images) of 23 individuals and reached an overall accuracy of 92.5\%.
\par
Finally, \cite{SIPEC_2020}~aims at providing behaviour tool analysis by offering segmentation, identification, pose-estimation and classification of behaviour from home-cage cameras. Well adapted, such a method may be resourceful for behaviour analysis in simple context~\cite{apes_mirror_2021} and help in the automatic acquisition and annotation of video recordings.
\par
This project aims to provide a pipeline for individual recognition of bonobos from a webcam located on an apparatus dedicated to data acquisition using a touch-screen. This work was inspired by the ZACI project~\cite{ZACI_Schmitt_2019}. Its overall goal is to automatise the data collection for cognitive studies in bonobos. This work differs from the previously mentioned projects by the complexity of the task and the dataset. The dataset is acquired automatically without single-image annotation which may lead to errors. Furthermore, the classification procedure may not focus on the face, but only on the visible parts of the body such as the back, top of the head or even just the limbs.
\par

In \autoref{sec:dataset} we present the context of the acquisition, annotation of the Bonobo dataset, and the different splitting methods carried out for evaluating the different classification methods presented in \autoref{sec:method}. The results of the classification methods on the different splits of the dataset are presented and discussed in \autoref{sec:Results}. Finally, we draw our conclusion and present the future planned work in \autoref{sec:conclusion}. 

\section{Weakly Annotated Bonobo Dataset}
\label{sec:dataset}

The Bonobo dataset, the classification models and the training methods presented are available on the Project GitHub page\footnote{\url{github.com/ccp-eva/BonobosClassification}}. In the following subsections, we describe the video acquisition, the weakly annotation procedure, the individual detector and the splitting methods of the dataset.

\subsection{Video Acquisition and Annotation}

\begin{figure}
    \centering
    \includegraphics[width=\linewidth]{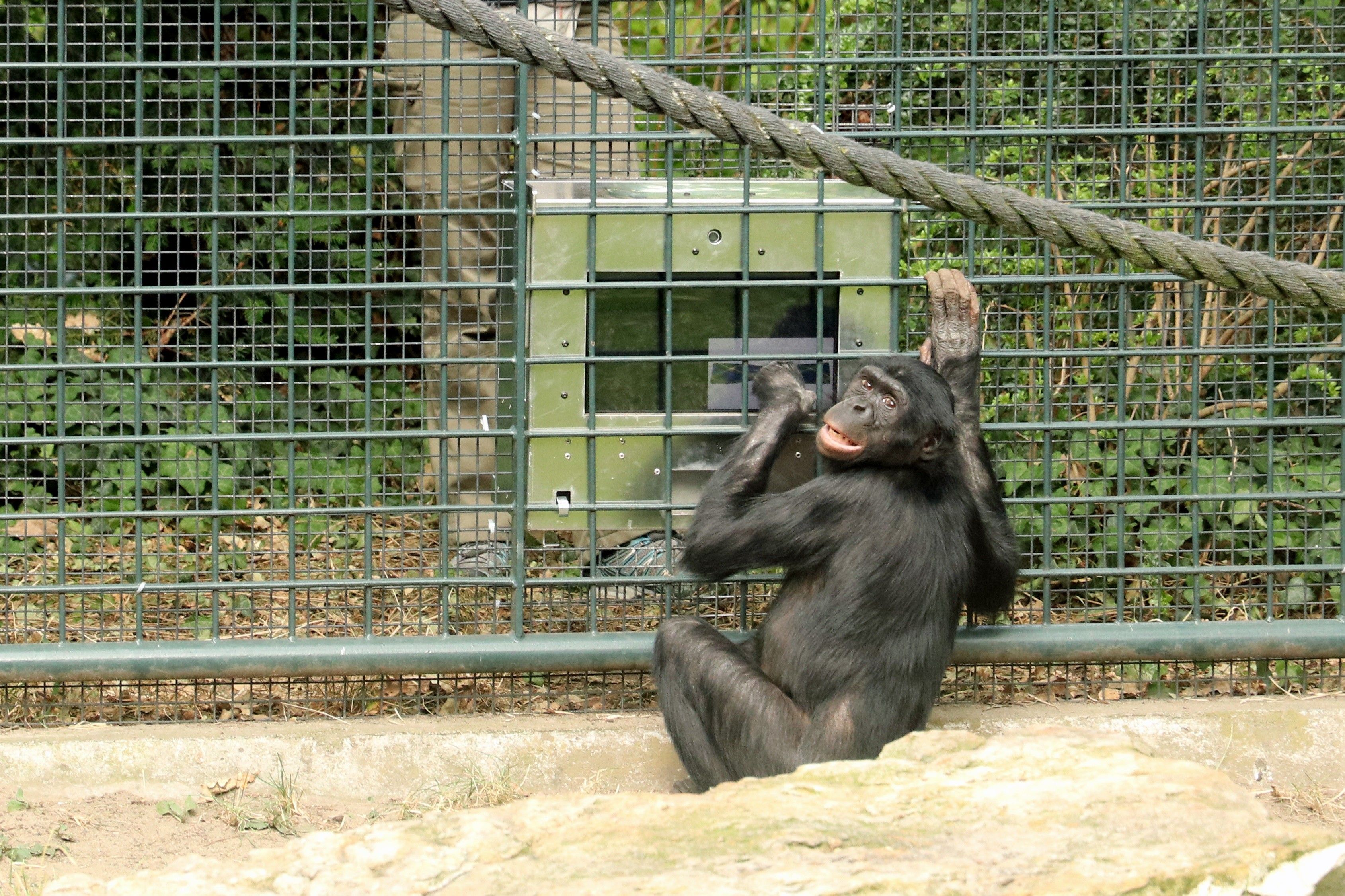}
    \caption{Video acquisition using the webcam located in the middle on the top side of the ZACI device. (\copyright 2018 Ruben Gralki/Zoo Berlin)}
    \label{fig:DataAcquisition}
\end{figure}

Videos were recorded using a digital camcorder Panasonic HC-V757 and a cheap Logitech webcam, both of resolution 1280x720 recording at 30~fps at the Zoo Berlin. The camcorder was manipulated by researchers familiar with the bonobos present in the zoo; while the webcam was located in the ZACI apparatus, see \autoref{fig:DataAcquisition}. The videos were then selected and sorted according to the individual present in the video. The videos can be assimilated to focal observation, a common practice in behaviour analysis that consists of observing (here filming) one particular individual and observing his/her actions and interactions. This may lead to having several individuals in the field of the camera, or none because of obstruction, camera manipulation, or in the eventuality of not having the individual in the webcam's field of view. Indeed, according to the bonobo position to the apparatus, the webcam may not be able to capture the individual, even when performing on the touch screen, as is the case on \autoref{fig:DataAcquisition}. No spatial information was annotated, nor is the presence of the individual in the field of the camera if several individuals were in the field of view. In this particular enclosure there are seven individuals  of different gender and age (gender/year of birth): Matayo (male/2019), Monyama (female/2010), Opala (female/1998), Santi (male/1981), Limbuko (male/1995), Leki (female/2014) and Samani female/2020). Samani was not incorporated into this first version of the dataset because of her constant proximity to her mother Monyama. A total of $100$ videos inequitably distributed across six bonobo individuals is here considered.

\subsection{Bonobo Detection}

OpenMMLab provides a Topdown Heatmap using a cascade RCNN x101~\cite{rcnn_x101_WangSCJDZLMTWLX19} coupled with an HRNet~\cite{HRNET_Sun_2019_CVPR} trained on MacaquePose~\cite{MacaquePose_Labuguen2021} in order to detect macaques and estimate their pose from images and videos. We use the output of the pre-trained RCNN x101 to assess the bonobo's location in the video. Even if bonobos and macaques are different in many aspects, the RCNN model can detect bonobos from the recorded videos. If several bonobos are detected in one frame, only the detection with the highest score is considered. Indeed, we assume focal observation of one individual would better capture this particular individual, correlating with its confidence score on the detector. This way, we can build a weakly labelled dataset based on videos associated with individuals and provide a frame-wise Region Of Interest (ROI) and a confidence score per video.

\subsection{Data Splitting}
\subsubsection{Splitting according to detection}

The early results in detection and classification, described in \autoref{sec:Results}, motivated a particular splitting of the obtained dataset. From it, four datasets are generated: \textit{noROI,SO}, \textit{ROI,SO}, \textit{noROI,SO.5} and \textit{ROI,SO.5}. They take into account two criteria: ROI and score. The generated dataset will either consider the ROI (and marked with \textit{ROI}) or not (\textit{noROI}) and consider the detection regardless of the score (\textit{S0}) or considering only detected bonobos with a score above or equal to 0.5 ((\textit{S0.5}). By considering only a better score, the number of samples per individual is impacted, making the classification task harder; but the quality of the data may be better, which may conversely ease the task. A score higher than 0.5 would lead to videos without detection and we, therefore, consider only these two generated datasets. Indeed, the score certainly remains low because of the difference between macaques and bonobos.

\par

In total, regardless of the score, $84$~$841$ bonobo detections are considered against $54~345$, with a score above 0.5, from the $100$ videos corresponding to $129~334$ frames. As presented in \autoref{tab:datasets}, compared to other similar datasets, the generated ones stand out by their number of samples per individual.

\begin{table}
    \small 
	\centering
		\begin{tabular}{|cccc|}
		\hline
  Dataset &  \# Samples & \# Classes & \# [min,max]/id \\ \hline
  Rhesus Macaques   & $679$     & $93$    & $[4, 192]$ \\
  C-Zoo             & $72~109$  & $24$    & $[62, 111]$ \\
  C-Tai             & $5$~$057$ & $66$    & $[4, 416]$  \\ \hline
  Bonobos S0        & $84~841$  & $6$     & $[2~590, 29~933]$  \\ 
  Bonobos S0.5      & $54~345$  & $6$     & $[2~214, 19~486]$  \\ \hline
    \end{tabular}
	\caption{Summary of similar datasets (inspired from~\cite{PrimateFaceIdentification_2019}) incorporating our generated datasets. The numbers in brackets show the range of samples per individual highlighting the imbalance of the dataset.}
	\label{tab:datasets}
 \end{table}

\subsubsection{Splitting According to Videos per Individual}

The dataset is also split following the proportion of videos 0.6, 0.2, and 0.2 respectively in train, validation and test sets for the videos associated per individual. This split is fundamental to having non-similar images across the splits. The number of frames per video was not considered for this splitting method, mainly because it changes according to the detection score. The split is performed once and randomly.
\par
\begin{figure}
    \centering
    \includegraphics[width=.9\linewidth]{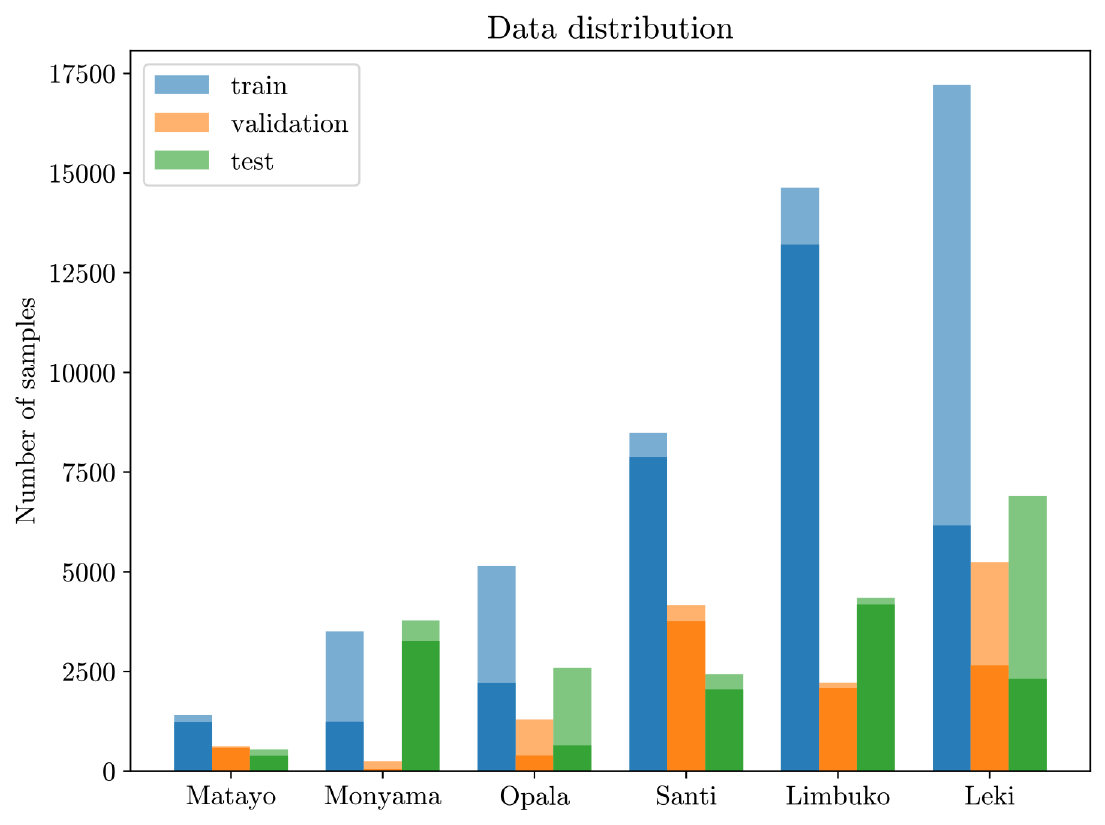}
    \caption{Data distribution across train, validation and test sets. Light and hard colors representing respectively \textit{S0} and \textit{S0.5} data distribution.}
    \label{fig:DatasetDist}
\end{figure}

The distribution of the samples per individual across train, validation and test for \textit{S0} and \textit{S0.5} is depicted in \autoref{fig:DatasetDist}.

\section{Bonobo Classification Methods}

\label{sec:method}

In this work, we are interested in comparing the commonly used machine learning classifier based on handcrafted features and the deep learning models as feature extractors or fine-tuned using pre-trained weights.

\subsection{Machine Learning Classifiers}

In order to perform classification, we consider seven machine learning algorithms: Logistic Regression (LR), Linear Discriminant Analysis (LDA), Gaussian Naive Bayes (NB), K-nearest neighbours (KNN), Support Vector Machine (SVM), Classification and Regression Trees (CART) and Random Forest (RF). All classifiers use the same basic image feature descriptors concatenated together: hu moments, haralick texture and color histogram (8 bins) computed on the images. These classifiers were trained on the whole generated datasets using a 10-fold cross-validation method. After obtaining these first results, we decided to split the dataset into train, validation, and test sets.

\subsection{Deep Learning Classifiers}

We compared the previous classifiers with the pre-trained ResNet18 model from~\cite{Resnet18_CVPR2016}. We run experiments only on the split datasets for this model. The model was used as a feature extractor: only the last fully-connected layer is trained. We also fine-tuned the model: all layers are trained. The same training method is used for the two cases for training: we train over 100 epochs with a decreased learning of factor 0.1 starting at 0.001 every 20 epochs. The model is fed with a batch of size 64 following a summing reduction to avoid overestimated backpropagation on shorter batches. We tried training with two losses: the usual cross-entropy loss and a weighed cross-entropy loss. The weighed cross-entropy loss uses the appearance of the individual in the training set: 
\begin{equation}
l_n = - w_{y_n} \log \frac{\exp(x_{n,y_n})}{\sum_{c=1}^C \exp(x_{n,c})}
\end{equation}

with $w_y=N_{classes}*\frac{N_{y}}{N}$. $N_{classes}$ represents the number of considered classes in our classification problem (6), $N_{y}$ the number of samples for a particular individual, and $N$ the total number of samples in the training set. This weight would equal $1$ if the samples were distributed evenly across individuals on the training set. Simple data augmentation is performed through random resizing, cropping and flipping within the scale $[0.08, 1]$ and ration $[\frac{3}{4}, \frac{4}{3}]$ and flipping probability of 0.5. These augmentation parameters are the same ones used for training the inception modules~\cite{NN:GoogLeNetInception,NN:InceptionV4_ResNet}. The images of all sets are then resized to $224\times224$ to fit the original input size of the pre-trained model. The model's state performing the best on the validation set with regard to the classification accuracy is saved for evaluation on the test set.

\section{Results and Analysis}
\label{sec:Results}

\subsection{Detection Results}

\begin{figure}
    \label{fig:detections}
    \begin{minipage}[b]{.48\linewidth}
      \centering
      \centerline{\includegraphics[width=\linewidth]{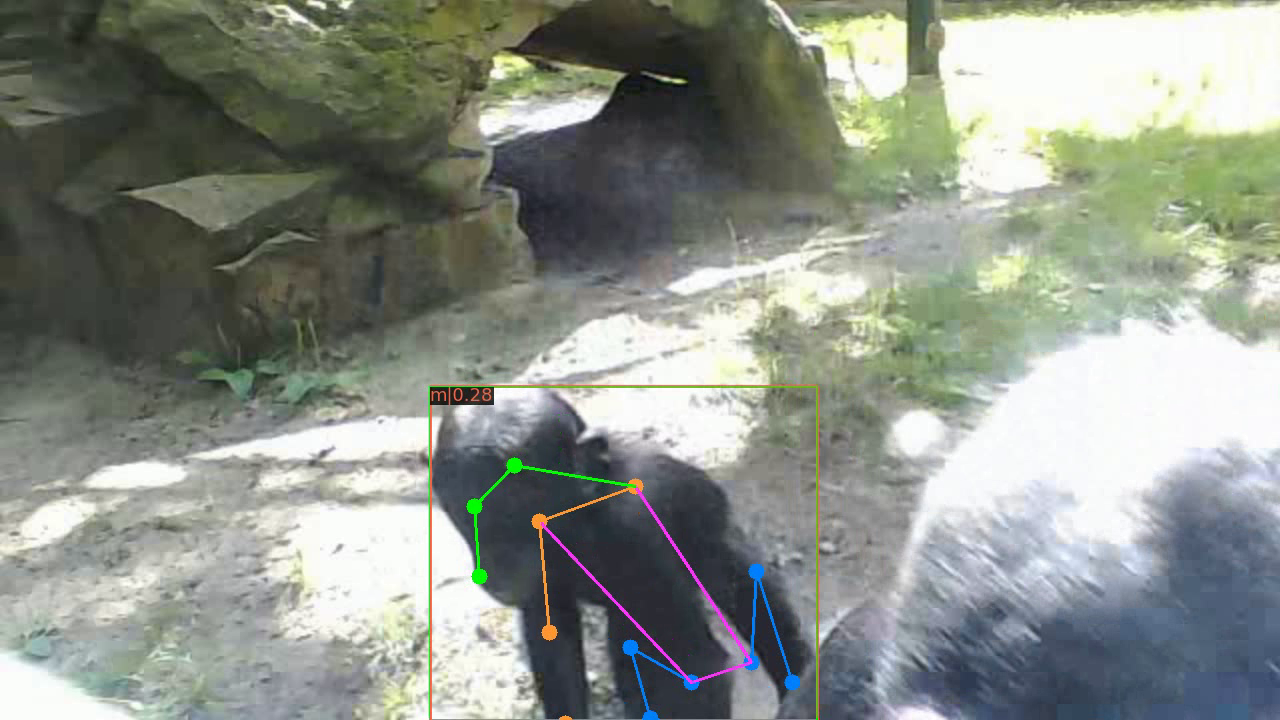}}
      \centerline{\textbf{a.} Opala and Matayo from webcam}\medskip
    \end{minipage}
    \begin{minipage}[b]{.48\linewidth}
      \centering
      \centerline{\includegraphics[width=\linewidth]{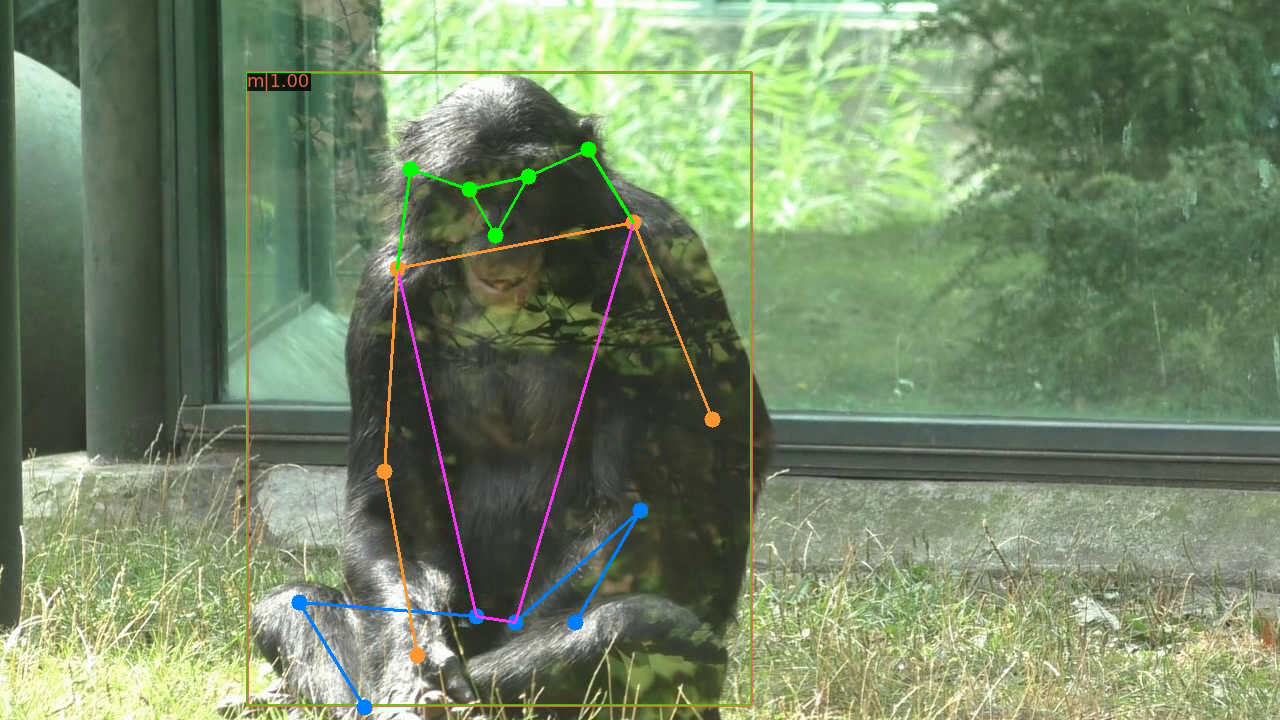}}
      \centerline{\textbf{b.} Limbuko from camcorder}\medskip
    \end{minipage}
    \caption{Bonobo detection results from the webcam in ZACI device and camcorder from public area in Zoo Berlin.}
\end{figure}

The ROI computed from the videos was automatically generated using a pre-trained method without the ground-truth at our disposal. Thus we can only have a limited appreciation of the detection results. As depicted in \autoref{fig:detections}, we can point out the limitation of our method for generating our different datasets. Indeed, in \autoref{fig:detections}.a, we may notice that the bonobo performing on the ZACI device is Opala (therefore a video weakly annotated as Opala), but it is instead her son Matayo which has been detected with a low score of 0.28. However, in \autoref{fig:detections}.b, we can notice the effectiveness of the detector on stable images from the camcorder despite a reflection due to the filming condition (behind a glass window from the public area of the Zoo of Berlin).

\subsection{Classification Results}

According to the cross-validation evaluation results reported in table \autoref{table:CrossVal}, we could think the classification methods are performing brilliantly and that the classification problem was easy to solve. Indeed, we have almost 100\% classification accuracy on all the generated datasets, especially with the Random Forest classifier. However, \autoref{table:Accs} shows us that such affirmation is inaccurate since performances are dropping for the same methods on the validation and test sets. Such behaviour can be explained by the high similarity of the data extracted from the videos while performing 10-fold cross-validation. Indeed, no separation of the data was performed concerning the videos for this evaluation method.

\begin{table}
    \small
	\centering
	\caption{Model accuracies* comparison with regard to the different generated datasets with 10-fold-cross validation method on the whole dataset.}
	\label{table:CrossVal}
	\begin{tabular}{|c|cccc|}
		\cline{2-5}
        \multicolumn{1}{c|}{}       & \multicolumn{4}{c|}{Generated Datasets} \\ \hline
        Models  & noROI,S0 & ROI,S0   & noROI,S0.5   & ROI,S0.5 \\ \hline
		LR      & .989  & .972   &  .995 & .993   \\ 
		LDA     & .971  & .959   &  .987 & .983   \\ 
		NB      & .633  & .614   &  .709 & .674   \\ 
		KNN     & .9995 & .998   &  .9993 & .9989  \\ 
		SVM     & .993  & .986   &  .997 & .997   \\
		CART    & .998  & .992   &  .998 & .996   \\ 
		RF      & \bf.9999 & \bf.999   &  \bf.9998 & \bf.9996  \\ \hline
	\end{tabular}
	\\[1pt] \raggedright
	*: the precision is set to 3 digits but may increase for better comparison.
\end{table}

\par
Results reported on \autoref{table:Accs} are less extreme but more complicated to interpret. It stresses the complexity and the variety of the different generated datasets. That is why we also give the mean accuracy across all classes on the test set (\textit{avgT}). The standard deviation is not reported but is below $10^2$. Are in bold font the best accuracies for validation test, and avgT. Globally, the convergence of the trained models was observed within the first 15 epochs.

\begin{table}
	\centering
	\caption{Model accuracies comparison with regard to the different generated datasets. The best values in the non-train sets are in bold per model type.}
	\label{table:Accs}
	\begin{tabular}{|c|cccc|cccc|cccc|cccc|}
		\cline{2-17}
        \multicolumn{1}{c|}{}       & \multicolumn{16}{c|}{Generated Datasets} \\ \cline{2-17}
        \multicolumn{1}{c|}{}       & \multicolumn{4}{c|}{noROI, S0}  & \multicolumn{4}{c|}{ROI, S0}    & \multicolumn{4}{c|}{noROI, S0.5}    & \multicolumn{4}{c|}{ROI, S0.5}  \\ \hline
		Models  & Tr. & Val.& T.  & mT. & Tr. & Val.& T.  & mT. & Tr. & Val.& T.  & avgT & Tr. & Val.& T.  & mT. \\ \hline
		LR                   & .99 & .38 & .34 & .17 & .98 & .38 & .34 & .17 & $1$ & .28 & .18 & .17 & .99 & .28 & .18 & .17 \\ 
		LDA                  & .98 & .38 & .34 & .17 & .97 & .30 & .12 & .17 & .99 & .28 & .18 & .17 & .99 & .28 & .18 & .17 \\ 
		NB                   & .68 & .38 & .34 & .17 & .63 & .38 & .34 & .17 & .80 & .28 & .18 & .17 & .78 & .28 & .18 & .17 \\
		KNN                  & $1$ & .38 & .34 & .17 & $1$ & .38 & .34 & .17 & $1$ & .28 & .18 & .17 & $1$ & .28 & .18 & .17 \\ 
		SVM                  & .99 & .38 & .34 & .17 & .99 & .38 & .17 & .17 & $1$ & .28 & .18 & .17 & $1$ & .28 &  .18 & .17 \\
		CART                 & $1$ & .43 & .61 & \bf.60 & $1$ & .33 & .41 & .32 & $1$ & .32 & .42 & \bf.45 & $1$ & .20 & .23 & .37 \\ 
		RF                   & $1$ & \bf.55 & \bf.65 & .49 & .99 & \bf.56 & \bf.66 & \bf.49 & $1$ & \bf.37 & \bf.50 & .42 & $1$ & \bf.44 & \bf.52 & \bf.47 \\ \hline
		ResNet               & .93 & \bf.85 & \bf.64 & \bf.49 & .91 & .\bf79 & .62 & .55 & .95 & \bf.76 & .49 & \bf.46 & .97 & .71 & .48 & .53  \\ 
		ResNet*              & .90 & .85 & .60 & .47 & .87 & .77 & .60 & .54 & .94 & .70 & .49 & .45 & .92 & \bf.75 & .48 & \bf.54  \\ 
		ResNet${\dagger}$    & .86 & .79 & .64 & .48 & .96 & .60 & \bf.75 & \bf.63 & .82 & .71 & \bf.50 & .45 & .64 & .49 & \bf.50 & .34  \\ 
		ResNet*${\dagger}$   & .99 & .67 & .61 & .45 & .93 & .54 & .58 & .46 & .99 & .47 & .39 & .39 & .83 & .41 & .42 & .41  \\ \hline
% 		ResNet*${\dagger}$      & .99   &  .56 & .52    & .26  & .54   &  .98 & .52    & .45  & .99   &  .75 & .78    & .89  & .48   &  .85 & .96    & .99  \\ \hline
	\end{tabular}
	\\[1pt] * : with weighted loss \hspace{1cm} ${\dagger}$ : fine-tuned
\end{table}

\par
The best performance using handcrafted features is obtained with the RF method. They perform correctly compared to the ResNet models. Still, ResNet generally outperforms the RF methods. The feature extractor using classical cross-entropy loss seems the best suited to classify our noROI datasets. Compared to the RF method, they hold similar test accuracy but have much higher validation accuracy. This may be explained by the ability of the model to extract meaningful local features from bigger images.

\par
The feature extractor trained using the weighted loss is the model performing the best on what we may consider the cleanest dataset (ROI with a confidence score above 0.5 - \textit{ROI, S0.5}). It gets the highest validation and test average accuracies while its accuracy on the test set (48\%) remains close to the best accuracy obtained by the fine-tuned ResNet (50\%). Surprisingly the weighted loss did not help much to improve the average test accuracy.

\par
\begin{figure}
    \label{fig:confmatrix}
    \centering
    \includegraphics[width=.95\linewidth]{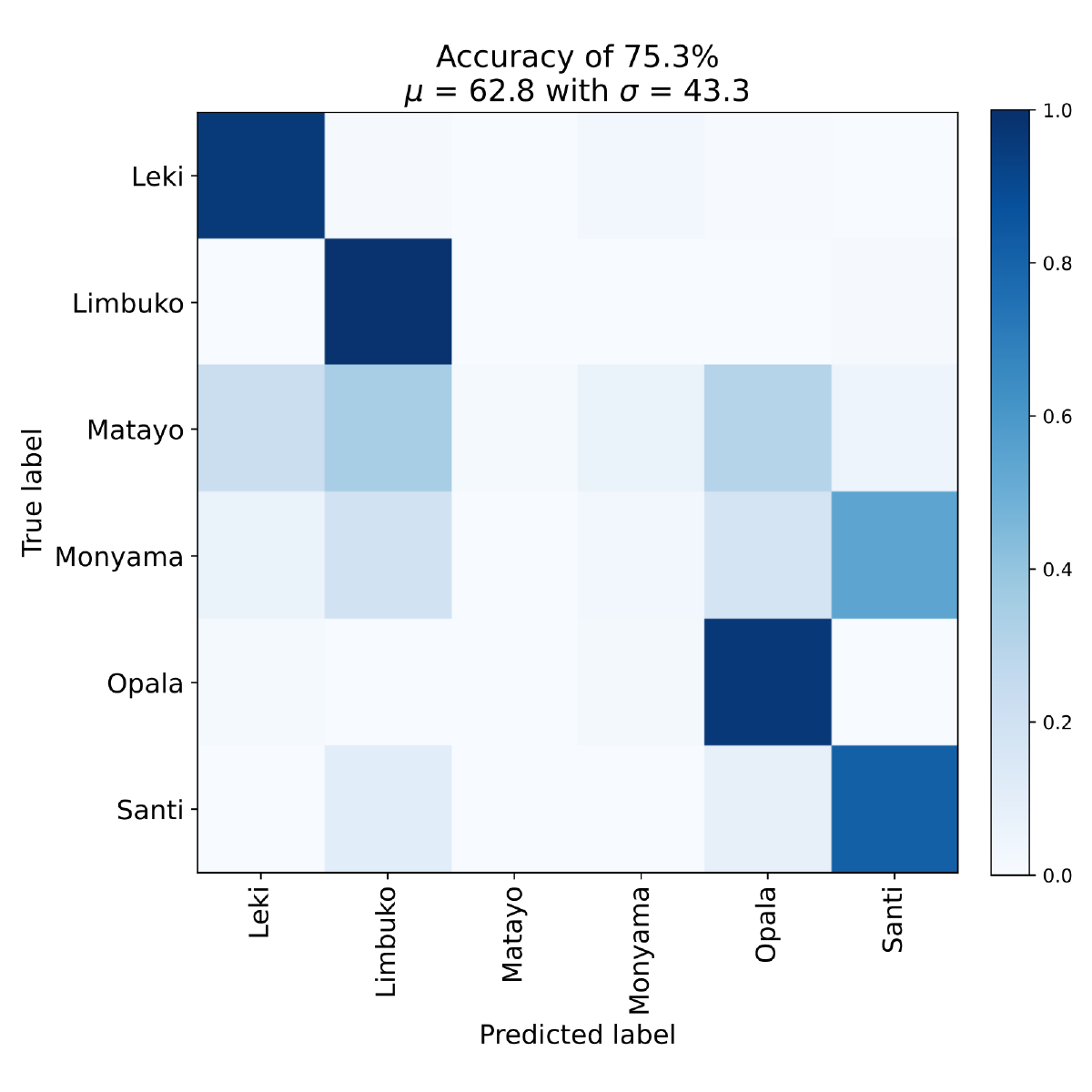}
    \caption{Confusion Matrix of fined-tuned ResNet model on the \textit{ROI, S0} dataset using non-weighted cross-entropy loss.}
\end{figure}

Furthermore, the fined-tuned ResNet on the \textit{ROI, S0} dataset is the one performing the best and getting the highest test and test average accuracy across all datasets and models. Its confusion matrix on the test set is depicted in \autoref{fig:confmatrix}. The model seems to have captured more discriminant information by updating the convolutional layers' weights with all detected ROI to solve this classification task. Still, Matayo and Monyama whose number of samples is the lowest in the dataset are without surprise the hardest to recognise. We may also notice the performance gap with the validation set underlying the difference between the train, validation and test sets and the difficulty of the task through the separation of the data. 

\par
Finally, we observed similar performance when using a fixed size of images for feature extraction. Similar results were obtained when saving the model state with regard to the lowest loss. Also, we observed a general tendency of the models to classify individuals as Leki or Santi, undoubtedly because of their higher appearance rate in the training set. It has been less observed with Limbuko when using the entire frame for classification. Indeed despite his high concentration in the train set, most of his videos were recorded using the camcorder. When using the whole frame, the classifier may recognise the scene and acquisition condition, and may not focus on the individual to classify. The integration of negative samples in the dataset did not help in the classification process.

\par
Overall, classification scores remain lower compared to other similar datasets because the region of interest used for classification is more challenging to classify, even for our team familiar with the bonobos. This may be overcome by considering the temporal information of the videos, building a track and classifying it. The several appearances of the individual in the track should help in the classification decision.

\section{Conclusion and Future Work}
\label{sec:conclusion}

This paper offers a pipeline to generate rich datasets of bonobos individuals using pre-trained detection models and a weak annotation method. We show the importance of video separation when performing a classification task to avoid false high hopes in classification performance. We investigate the most common classification methods in machine learning to solve the bonobo individual classification and show the superiority of deep-learned features compared to handcrafted features. The ResNet performance is analysed and discussed using different training methods on the different generated datasets. To achieve better performance, we believe more complex network architecture is not the priority, instead, we wish to deepen the analysis of the detection performance and the quality of the generated datasets.
\par 
Future work will focus on the annotation of the acquired videos using CVAT~\cite{cvat} tool in order to evaluate the detection method and the automatically built datasets qualities. Other classification methods will also be investigated using direct feature differences to estimate the similarity between individuals and allow more flexible applications in different zoos to perform classification on a larger number of individuals without retraining the model. The focus will also be on using temporal information for classification, tracking and activity recognition in primates for further behaviour analysis.

\subsubsection{Acknowledgements}
Thanks to Zoo Berlin and its responsible curator, Dr.~André Schüle; its deputy head keeper and bonobo keeper, Ruben Gralki and Kathrin Susanne Kopp from MPI EVA for the acquisition and annotation of the Bonobo dataset.
\newpage

\bibliographystyle{bibtex/spmpsci}
\bibliography{refs}

\begin{thebibliography}{10}
\providecommand{\url}[1]{{#1}}
\providecommand{\urlprefix}{URL }
\expandafter\ifx\csname urlstyle\endcsname\relax
  \providecommand{\doi}[1]{DOI~\discretionary{}{}{}#1}\else
  \providecommand{\doi}{DOI~\discretionary{}{}{}\begingroup
  \urlstyle{rm}\Url}\fi

\bibitem{apes_mirror_2021}
Cebioğlu, S., Broesch, T.: Explaining cross-cultural variation in mirror
  self-recognition: {New} insights into the ontogeny of objective
  self-awareness.
\newblock Developmental Psychology \textbf{57}(5), 625 (2021).
\newblock \doi{10.1037/dev0001171}.
\newblock \urlprefix\url{https://psycnet.apa.org/fulltext/2021-58449-002.pdf}.
\newblock Publisher: US: American Psychological Association

\bibitem{mmdetection_2019}
Chen, K., Wang, J., Pang, J., Cao, Y., Xiong, Y., Li, X., Sun, S., Feng, W.,
  Liu, Z., Xu, J., Zhang, Z., Cheng, D., Zhu, C., Cheng, T., Zhao, Q., Li, B.,
  Lu, X., Zhu, R., Wu, Y., Dai, J., Wang, J., Shi, J., Ouyang, W., Loy, C.C.,
  Lin, D.: {MMDetection}: Open mmlab detection toolbox and benchmark.
\newblock arXiv preprint arXiv:1906.07155  (2019)

\bibitem{CNNChimpReco_2016}
Freytag, A., Rodner, E., Simon, M., Loos, A., Kühl, H.S., Denzler, J.:
  Chimpanzee {Faces} in the {Wild}: {Log}-{Euclidean} {CNNs} for {Predicting}
  {Identities} and {Attributes} of {Primates}.
\newblock In: B.~Rosenhahn, B.~Andres (eds.) Pattern {Recognition}, Lecture
  {Notes} in {Computer} {Science}, pp. 51--63. Springer International
  Publishing, Cham (2016).
\newblock \doi{10.1007/978-3-319-45886-1_5}

\bibitem{Animal_Face_Primates_2020}
Guo, S., Xu, P., Miao, Q., Shao, G., Chapman, C.A., Chen, X., He, G., Fang, D.,
  Zhang, H., Sun, Y., Shi, Z., Li, B.: Automatic {Identification} of
  {Individual} {Primates} with {Deep} {Learning} {Techniques}.
\newblock iScience \textbf{23}(8), 101,412 (2020).
\newblock \doi{10.1016/j.isci.2020.101412}.
\newblock
  \urlprefix\url{https://www.sciencedirect.com/science/article/pii/S2589004220306027}

\bibitem{Resnet18_CVPR2016}
He, K., Zhang, X., Ren, S., Sun, J.: Deep residual learning for image
  recognition.
\newblock In: 2016 {IEEE} Conference on Computer Vision and Pattern
  Recognition, {CVPR} 2016, Las Vegas, NV, USA, June 27-30, 2016, pp. 770--778.
  {IEEE} Computer Society (2016).
\newblock \doi{10.1109/CVPR.2016.90}.
\newblock \urlprefix\url{https://doi.org/10.1109/CVPR.2016.90}

\bibitem{pandaidentification_2020}
Hou, J., He, Y., Yang, H., Connor, T., Gao, J., Wang, Y., Zeng, Y., Zhang, J.,
  Huang, J., Zheng, B., Zhou, S.: Identification of animal individuals using
  deep learning: A case study of giant panda.
\newblock Biological Conservation \textbf{242}, 108,414 (2020).
\newblock \doi{https://doi.org/10.1016/j.biocon.2020.108414}.
\newblock
  \urlprefix\url{https://www.sciencedirect.com/science/article/pii/S000632071931609X}

\bibitem{Animal_biometric_2017}
Kumar, S., Singh, S.K., Singh, R., Singh, A.K.: Animal Biometrics: Concepts and
  Recent Application, pp. 1--20.
\newblock Springer Singapore, Singapore (2017).
\newblock \doi{10.1007/978-981-10-7956-6_1}.
\newblock \urlprefix\url{https://doi.org/10.1007/978-981-10-7956-6_1}

\bibitem{MacaquePose_Labuguen2021}
Labuguen, R., Matsumoto, J., Negrete, S.B., Nishimaru, H., Nishijo, H., Takada,
  M., Go, Y., Inoue, K.I., Shibata, T.: Macaquepose: A novel "in the wild"
  macaque monkey pose dataset for markerless motion capture.
\newblock Frontiers in behavioral neuroscience \textbf{14}, 581,154--581,154
  (2021).
\newblock \doi{10.3389/fnbeh.2020.581154}.
\newblock \urlprefix\url{https://pubmed.ncbi.nlm.nih.gov/33584214}.
\newblock 33584214[pmid]

\bibitem{ChimpdetectandidentifSURFGABOR_2013}
Loos, A., Ernst, A.: An automated chimpanzee identification system using face
  detection and recognition.
\newblock EURASIP Journal on Image and Video Processing \textbf{2013}(1), 49
  (2013).
\newblock \doi{10.1186/1687-5281-2013-49}.
\newblock \urlprefix\url{https://doi.org/10.1186/1687-5281-2013-49}

\bibitem{Primate_id_handcraftedfeatures_2012}
Loos, A., Pfitzer, M.: Towards automated visual identification of primates
  using face recognition.
\newblock In: 2012 19th {International} {Conference} on {Systems}, {Signals}
  and {Image} {Processing} ({IWSSIP}), pp. 425--428 (2012).
\newblock ISSN: 2157-8702

\bibitem{SIPEC_2020}
Marks, M., Qiuhan, J., Sturman, O., von Ziegler, L., Kollmorgen, S., von~der
  Behrens, W., Mante, V., Bohacek, J., Yanik, M.F.: Sipec: the deep-learning
  swiss knife for behavioral data analysis.
\newblock bioRxiv  (2020).
\newblock \doi{10.1101/2020.10.26.355115}.
\newblock
  \urlprefix\url{https://www.biorxiv.org/content/early/2020/10/29/2020.10.26.355115}

\bibitem{Deeplabcut_Mathisetal2018}
Mathis, A., Mamidanna, P., Cury, K.M., Abe, T., Murthy, V.N., Mathis, M.W.,
  Bethge, M.: Deeplabcut: markerless pose estimation of user-defined body parts
  with deep learning.
\newblock Nature Neuroscience  (2018).
\newblock \urlprefix\url{https://www.nature.com/articles/s41593-018-0209-y}

\bibitem{ScikitLearn_2011}
Pedregosa, F., Varoquaux, G., Gramfort, A., Michel, V., Thirion, B., Grisel,
  O., Blondel, M., Prettenhofer, P., Weiss, R., Dubourg, V., et~al.:
  Scikit-learn: Machine learning in python.
\newblock Journal of machine learning research \textbf{12}(Oct), 2825--2830
  (2011)

\bibitem{ZACI_Schmitt_2019}
Schmitt, V.: Implementing portable touchscreen-setups to enhance cognitive
  research and enrich zoo-housed animals.
\newblock Journal of Zoo and Aquarium Research \textbf{7}(2), 50–58 (2019).
\newblock \doi{10.19227/jzar.v7i2.314}.
\newblock \urlprefix\url{https://jzar.org/jzar/article/view/314}

\bibitem{CNN_Chimp_recognition_2019}
Schofield, D., Nagrani, A., Zisserman, A., Hayashi, M., Matsuzawa, T., Biro,
  D., Carvalho, S.: Chimpanzee face recognition from videos in the wild using
  deep learning.
\newblock Science Advances \textbf{5}(9), eaaw0736 (2019).
\newblock \doi{10.1126/sciadv.aaw0736}.
\newblock
  \urlprefix\url{https://www.science.org/doi/full/10.1126/sciadv.aaw0736}.
\newblock Publisher: American Association for the Advancement of Science

\bibitem{cvat}
Sekachev, B., Manovich, N., Zhiltsov, M., Zhavoronkov, A., Kalinin, D., Hoff,
  B., TOsmanov, Kruchinin, D., Zankevich, A., DmitriySidnev, Markelov, M.,
  Johannes222, Chenuet, M., a~andre, telenachos, Melnikov, A., Kim, J., Ilouz,
  L., Glazov, N., Priya4607, Tehrani, R., Jeong, S., Skubriev, V., Yonekura,
  S., vugia truong, zliang7, lizhming, Truong, T.: opencv/cvat: v1.1.0 (2020).
\newblock \doi{10.5281/zenodo.4009388}.
\newblock \urlprefix\url{https://doi.org/10.5281/zenodo.4009388}

\bibitem{PrimateFaceIdentification_2019}
Shukla, A., Cheema, G.S., Anand, S., Qureshi, Q., Jhala, Y.: Primate {Face}
  {Identification} in the {Wild}.
\newblock In: A.C. Nayak, A.~Sharma (eds.) {PRICAI} 2019: {Trends} in
  {Artificial} {Intelligence}, Lecture {Notes} in {Computer} {Science}, pp.
  387--401. Springer International Publishing, Cham (2019).
\newblock \doi{10.1007/978-3-030-29894-4_32}

\bibitem{HRNET_Sun_2019_CVPR}
Sun, K., Xiao, B., Liu, D., Wang, J.: Deep high-resolution representation
  learning for human pose estimation.
\newblock In: Proceedings of the IEEE/CVF Conference on Computer Vision and
  Pattern Recognition (CVPR) (2019)

\bibitem{NN:InceptionV4_ResNet}
Szegedy, C., Ioffe, S., Vanhoucke, V., Alemi, A.A.: Inception-v4,
  inception-resnet and the impact of residual connections on learning.
\newblock In: S.P. Singh, S.~Markovitch (eds.) Proceedings of the Thirty-First
  {AAAI} Conference on Artificial Intelligence, February 4-9, 2017, San
  Francisco, California, {USA}, pp. 4278--4284. {AAAI} Press (2017).
\newblock
  \urlprefix\url{http://aaai.org/ocs/index.php/AAAI/AAAI17/paper/view/14806}

\bibitem{NN:GoogLeNetInception}
Szegedy, C., Liu, W., Jia, Y., Sermanet, P., Reed, S.E., Anguelov, D., Erhan,
  D., Vanhoucke, V., Rabinovich, A.: Going deeper with convolutions.
\newblock In: {IEEE} Conference on Computer Vision and Pattern Recognition,
  {CVPR} 2015, Boston, MA, USA, June 7-12, 2015, pp. 1--9. {IEEE} Computer
  Society (2015).
\newblock \doi{10.1109/CVPR.2015.7298594}.
\newblock \urlprefix\url{https://doi.org/10.1109/CVPR.2015.7298594}

\bibitem{rcnn_x101_WangSCJDZLMTWLX19}
Wang, J., Sun, K., Cheng, T., Jiang, B., Deng, C., Zhao, Y., Liu, D., Mu, Y.,
  Tan, M., Wang, X., Liu, W., Xiao, B.: Deep high-resolution representation
  learning for visual recognition.
\newblock TPAMI  (2019)

\bibitem{Rhesus_macaques_2018}
Witham, C.L.: Automated face recognition of rhesus macaques.
\newblock Journal of Neuroscience Methods \textbf{300}, 157--165 (2018).
\newblock \doi{10.1016/j.jneumeth.2017.07.020}.
\newblock
  \urlprefix\url{https://www.sciencedirect.com/science/article/pii/S0165027017302637}

\bibitem{detectron2_facebook_2019}
Wu, Y., Kirillov, A., Massa, F., Lo, W.Y., Girshick, R.: Detectron2.
\newblock \url{https://github.com/facebookresearch/detectron2} (2019)

\end{thebibliography}

% %
% % ---- Bibliography ----
% %
% \begin{thebibliography}{6}
% %

% \bibitem {smit:wat}
% Smith, T.F., Waterman, M.S.: Identification of common molecular subsequences.
% J. Mol. Biol. 147, 195?197 (1981). \url{doi:10.1016/0022-2836(81)90087-5}

% \bibitem {may:ehr:stein}
% May, P., Ehrlich, H.-C., Steinke, T.: ZIB structure prediction pipeline:
% composing a complex biological workflow through web services.
% In: Nagel, W.E., Walter, W.V., Lehner, W. (eds.) Euro-Par 2006.
% LNCS, vol. 4128, pp. 1148?1158. Springer, Heidelberg (2006).
% \url{doi:10.1007/11823285_121}

% \bibitem {fost:kes}
% Foster, I., Kesselman, C.: The Grid: Blueprint for a New Computing Infrastructure.
% Morgan Kaufmann, San Francisco (1999)

% \bibitem {czaj:fitz}
% Czajkowski, K., Fitzgerald, S., Foster, I., Kesselman, C.: Grid information services
% for distributed resource sharing. In: 10th IEEE International Symposium
% on High Performance Distributed Computing, pp. 181?184. IEEE Press, New York (2001).
% \url{doi: 10.1109/HPDC.2001.945188}

% \bibitem {fo:kes:nic:tue}
% Foster, I., Kesselman, C., Nick, J., Tuecke, S.: The physiology of the grid: an open grid services architecture for distributed systems integration. Technical report, Global Grid
% Forum (2002)

% \bibitem {onlyurl}
% National Center for Biotechnology Information. \url{http://www.ncbi.nlm.nih.gov}

% \end{thebibliography}
\end{document}